\documentclass{article}
\usepackage{spconf}
\usepackage{amsmath,amsfonts}
\usepackage{algorithmic}
\usepackage{array}
\usepackage[caption=false,font=normalsize,labelfont=sf,textfont=sf]{subfig}
\usepackage{textcomp}
\usepackage{stfloats}
\usepackage{url}
\usepackage{verbatim}
\usepackage{graphicx}
\usepackage{placeins}
\usepackage{glossaries}
\usepackage{multirow}
\usepackage{multicol}
\usepackage{hyperref}
\usepackage[style=ieee]{biblatex}
\usepackage{balance}

\newacronym{bez}{BEZ}{Biometric Evaluation Center}
\newacronym{gdpr}{GDPR}{General Data Protection Regulation}
\newacronym{fr}{FR}{face recognition}
\newacronym{cots}{COTS}{commercial off-the-shelf}
\newacronym{incd}{INCD}{Israel’s National Cyber Directorate}
\newacronym{msu}{MSU}{Michigan State University}
\newacronym{hbrs}{H-BRS}{Bonn-Rhein-Sieg University}
\newacronym{bsi}{BSI}{Federal Office for Information Security}
\newacronym{abc}{ABC}{automated border control}
\newacronym{tft}{TFT}{thin film transistor}
\newacronym{tof}{ToF}{Time-of-Flight}
\newacronym{rppg}{rPPG}{remote photoplethysmography}
\newacronym{oct}{OCT}{optical coherence tomography}
\newacronym{nir}{NIR}{near-infrared}
\newacronym{pad}{PAD}{presentation attack detection}
\newacronym{far}{FAR}{False Acceptance Rate}
\newacronym{frr}{FRR}{False Rejection Rate}
\newacronym{eer}{EER}{Equal Error Rate}
\newacronym{fta}{FTA}{Failure to Acquire}
\newacronym{fmr}{FMR}{False Match Rate}
\newacronym{fnmr}{FNMR}{False Non-Match Rate}

\title{Longitudinal Study of Facial Biometrics at the BEZ: Temporal Variance Analysis}

\twoauthors
 {\shortstack{Mathias Schulz\sthanks{mathias.schulz@h-brs.de}, Pia Funk,\\
Markus Rohde, Norbert Jung, Robert Lange}}
    {Hochschule Bonn-Rhein-Sieg (H-BRS)\\
    Institute of Safety and Security Research\\
    Grantham-Allee 20\\
    D-53757 Sankt Augustin\\
    }
 {\shortstack{Alexander Spenke, Florian Blümel,\\
Ralph Breithaupt, Gerd Nolden}}
    {Federal Office for Information Security (BSI)\\
    Directorate General D Digitisation and Identities\\
    Godesberger Allee 87\\
    D-53175 Bonn}
    
\begin{document}
\maketitle

\begin{abstract}
This study presents findings from long-term biometric evaluations conducted at the \gls{bez}. Over the course of more than two and a half years, our ongoing research with over 400 participants representing diverse ethnicities, genders, and age groups were regularly assessed using a variety of biometric tools and techniques at the controlled testing facilities. Our findings are based on the  \gls{gdpr}- compliant local \gls{bez} database with more than 238,000 biometric data sets categorized into multiple biometric modalities such as face and finger.
We used state-of-the-art \gls{fr} algorithms to analyze long-term comparison scores. Our results show that these scores fluctuate more significantly between individual days than over the entire measurement period. These findings highlight the importance of testing biometric characteristics of the same individuals over a longer period of time in a controlled measurement environment and lays the groundwork for future advancements in biometric data analysis.
\end{abstract}


\section{Introduction}
\label{sec:introduction}
Obtaining statistically significant results has long been a challenge in biometric system evaluations. Traditional approaches often involved conducting single-session tests with as many participants as possible. However, this method made it difficult to identify meaningful statistical influences, as individual variances and uncontrolled conditions compromised the reliability and generalization of the results.

A long-term evaluation approach offers a solution to this problem by enabling systematic and controlled testing over extended periods. By maintaining controlled and adjustable conditions, several key advantages can be realized. First, the same systems can be tested with the same individuals over time, enabling robust longitudinal analysis, to evaluate the aging processes of people and systems. Second, different systems can be evaluated either simultaneously or within a short time frame under identical conditions, which facilitates direct comparisons. Third, newer versions of a system can be assessed in relation to their predecessors, providing insights into technological progress. In addition, extensions and modifications of existing systems, as well as the development of entirely new systems, can be evaluated in a structured manner. Finally, usability aspects can be examined thoroughly under these consistent conditions.

Despite the evident benefits of long-term evaluations, current research lacks extensive datasets that reflect such conditions. Most studies rely on short-term assessments, failing to capture the effects of continuous system use, adaptation, and potential performance fluctuations over time. This study aims to bridge this gap by establishing a comprehensive long-term evaluation framework that systematically examines system performance, usability, and user adaptation over an extended period.

\section{Related Work}
\label{sec:related_work}
\noindent In the field of biometric evaluations, traditional methodologies have predominantly relied on single-session data collection, where subjects are enrolled and authenticated within a limited time-frame. This approach, while practical, often fails to account for temporal variations in biometric traits, potentially impacting the long-term reliability and accuracy of biometric systems. However, a few research institutions worldwide addressed this issue by investigating the long-term stability of biometric traits and evaluating biometric systems over extended periods in their studies. These studies provide valuable insights into the effects of aging and repeated exposures on biometric performance.

The National Biometrics Laboratory at \gls{incd} has been instrumental in advancing biometric research, focusing on the long-term reliability of biometric systems. Their work contributes to national security measures and the development of robust biometric identification systems \cite{incd2025}.

In the United States, the Biometrics Research Group at \gls{msu} \cite{msubiometrics2025} has conducted extensive longitudinal studies on both fingerprint and facial recognition systems: 
A significant study by L. Best-Rowden and A. K. Jain analyzed the permanence of facial features over time \cite{best-rowden_longitudinal_2018}. Utilizing large-scale mugshot databases, the researchers applied mixed-effects regression models to assess the degradation in genuine similarity scores as the time between enrolled and query images increased. Their findings indicated a measurable decline in recognition accuracy over extended periods, underscoring the necessity for longitudinal data to enhance the robustness of facial recognition systems.  

Yoon and Jain conducted an extensive longitudinal analysis of fingerprint recognition, examining records from 15,597 subjects over a span of up to 12 years \cite{yoon_longitudinal_2015}. The study revealed a significant decrease in genuine match scores correlating with longer time intervals between fingerprint captures. Despite this decline, the overall recognition accuracy remained stable, provided the fingerprint images were of high quality. 

Both of these studies highlight the importance of considering temporal factors in biometrics to maintain system reliability. However, the temporal distribution of data in these studies is often highly variable, as they rely on operational datasets such as mugshots, resulting in inconsistent intervals between recordings. Consequently, while such datasets offer valuable insights into long-term biometric performance, they may lack the controlled conditions and consistent measurement intervals required for specific precise temporal analysis.

\section{Methodology}
\label{sec:methodology}
\FloatBarrier
\subsection{Participant acquisition}
\noindent This study was structured to allow repeated measurements of the same participants over an extended period to evaluate biometric characteristics under controlled long-term conditions. The measurements took place in the \gls{bez}, which is located directly on the \gls{hbrs} campus. Its central location proved to be highly advantageous for participant accessibility. Students and staff members were able to conveniently stop by during breaks, enabling frequent and flexible participation.

Building on this logistical advantage, a variety of recruitment strategies were employed to ensure a diverse and representative sample. Flyers were distributed across key locations such as cafeterias and faculty buildings. Additionally, university-wide emails were sent to inform students and staff about the study and encourage participation. Posters highlighting the study’s objectives and participation requirements were strategically placed in high-traffic areas, including hallways and department bulletin boards. To further raise awareness, brief presentations were given in selected lectures, providing students with a direct opportunity to ask questions and to learn about the study. Moreover, a promotional advertisement was displayed on the large LED screen above the main entrance of the \gls{hbrs}, ensuring high visibility to both students and faculty members. 

To create an additional incentive for participation in the evaluations, vouchers are issued to the subjects after several visits.

\subsection{BEZ evaluation process}
\noindent An additional important requirement, irrespective of the application, arises from the \gls{gdpr}: the processing of personal data inherent in all biometric procedures requires special measures to ensure sufficient data protection.

\begin{figure}[!h]
\centering
\includegraphics[width=\linewidth]{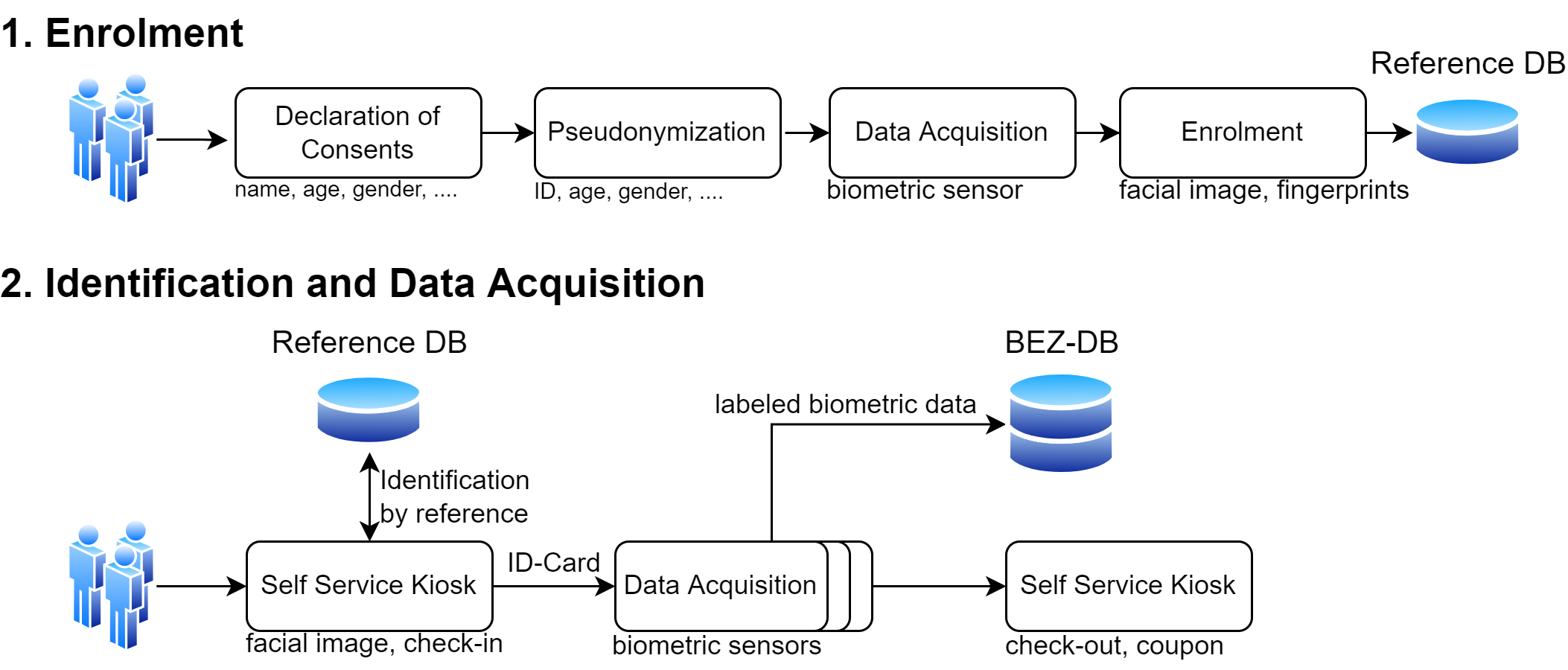}
\caption{\gls{gdpr} compliant \gls{bez} evaluation process. 1. Enrollment: declaration of consent and biometric reference data is recorded. 2. Identification: autonomous check-in via biometric reference and participant to evaluation. 
}
\label{fig:BEZ_Datenfluss_Diagramm}
\end{figure}

The biometric database used in this study was constructed based on the collected data from recruited participants under controlled conditions. The \gls{gdpr} compliant evaluation process is divided into two main phases, as illustrated in Fig.~\ref{fig:BEZ_Datenfluss_Diagramm}: First, each subject underwent an initial enrollment phase to establish reference biometric data and to familiarize themselves with the study procedures. At the beginning of the enrollment session, participants attended a brief introductory session, which included a short guided tour of the testing environment and a detailed explanation of the study’s objectives, methodology and privacy policies. Then participants signed an informed consent form, ensuring voluntary participation and compliance with ethical research guidelines. Within this process, every proband is issued with an unique ID to pseudonymize identities. Afterwards, access to the connection between ID and real name is only accessible for the \gls{bsi}’s data protection officer, who also administers the declarations of consents. Thus, the \gls{bez} databases only store the pseudonymized IDs along with the recorded data. The test subjects can at any time request information about their stored data as well as the deletion of this data, including any backups, from the data protection officer. Timely processing will then be ensured.

\begin{figure}[htbp]
\centering
\begin{minipage}[t]{0.3\textwidth}
\vspace{32pt}
\includegraphics[width=\linewidth]{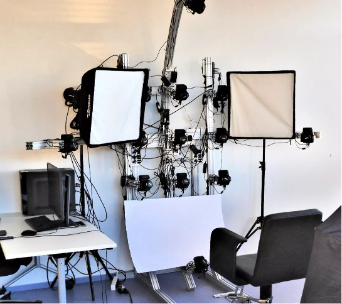}
\begin{center}
    (a)
\end{center}
\end{minipage}%
\begin{minipage}[t]{0.2\textwidth}
\vspace{0pt}
\includegraphics[width=\linewidth]{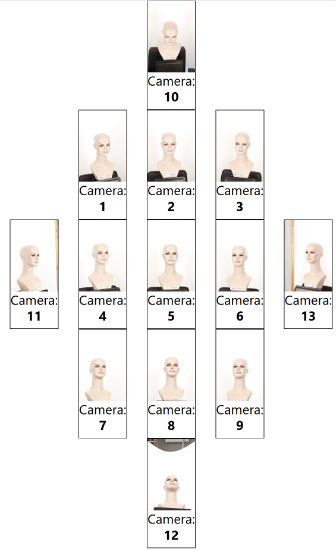}
\begin{center}
    (b)
\end{center}
\end{minipage}
\caption{(\textbf{a}) K13 photo studio setup for high quality images of 13 different angles.; (\textbf{b}) K13 camera angles setup. Camera 5 records a full frontal image whereas additional cameras are positioned in 15° and 45° degree angles both on horizontal and vertical axes.}
\label{fig:k13_experiment_setup}
\end{figure}

During the enrollment a high resolution and high quality set of facial images is captured with the K13 photo studio to serve as the initial reference for \gls{fr}. The K13 photo studio offers standardized lighting, uniform background and positional conditions. Each record includes high quality images from 13 cameras (see Fig.~\ref{fig:k13_experiment_setup}). With camera 5, a full frontal image of the face is captured. Eight additional cameras are positioned in angles of 15 degree, whereas the cameras on the diagonal have angles of 15° both on horizontal and vertical axes. The outer four cameras take images in 45° positions.

Additionally all fingers of the left and right hand are captured during the enrollment using a high-resolution optical four-finger fingerprint scanner. Each finger is captured in single finger capture mode.
These recorded reference facial image and fingerprints as well as the proband data (age, gender, etc.) and consents given are entered into the reference database. The stored reference data records of the test subjects are updated in a regular manner.

After enrollment, in the second phase the main data acquisition process is carried out. Probands can use the self-service kiosks for autonomous check in to the \gls{bez} evaluations. The kiosk first captures a facial image, which is then compared against the reference database of enrolled participants. In case of a single match, the subject is identified and the unique proband ID is found. The kiosk writes this ID together with the facial reference image from the database to a smartcard. This smartcard can then be used at the biometric data acquisition stations where proband ID is connected to all recordings of this person. Furthermore, due to the stored reference image, the smartcard can be used like a passport to enter the experimental \gls{abc} gates. The recorded biometric data together with experiment specific meta-information (e.g. subject ID, experiment, time, system-settings, environmental conditions, specific headgear and glasses...) is saved to the \gls{bez} database. To ensure easy handling and high-quality dataset curation, biometric samples of each experiment station is annotated by a custom labeling scheme. After evaluation, probands can go back to the self-services kiosk to check out, whereby all data on the used smartcard is automatically deleted. The subject receives a credit note for the participation.

\subsection{Datasets}
\begin{figure}[htbp]
\centering
\begin{minipage}[t]{0.25\textwidth}
\includegraphics[width=\linewidth]{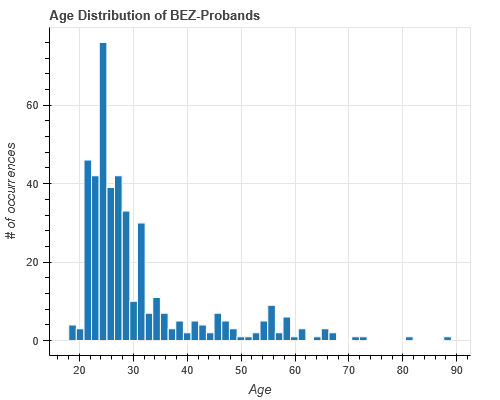}
\begin{center}
    (a)
\end{center}
\end{minipage}%
\begin{minipage}[t]{0.25\textwidth}
\includegraphics[width=\linewidth]{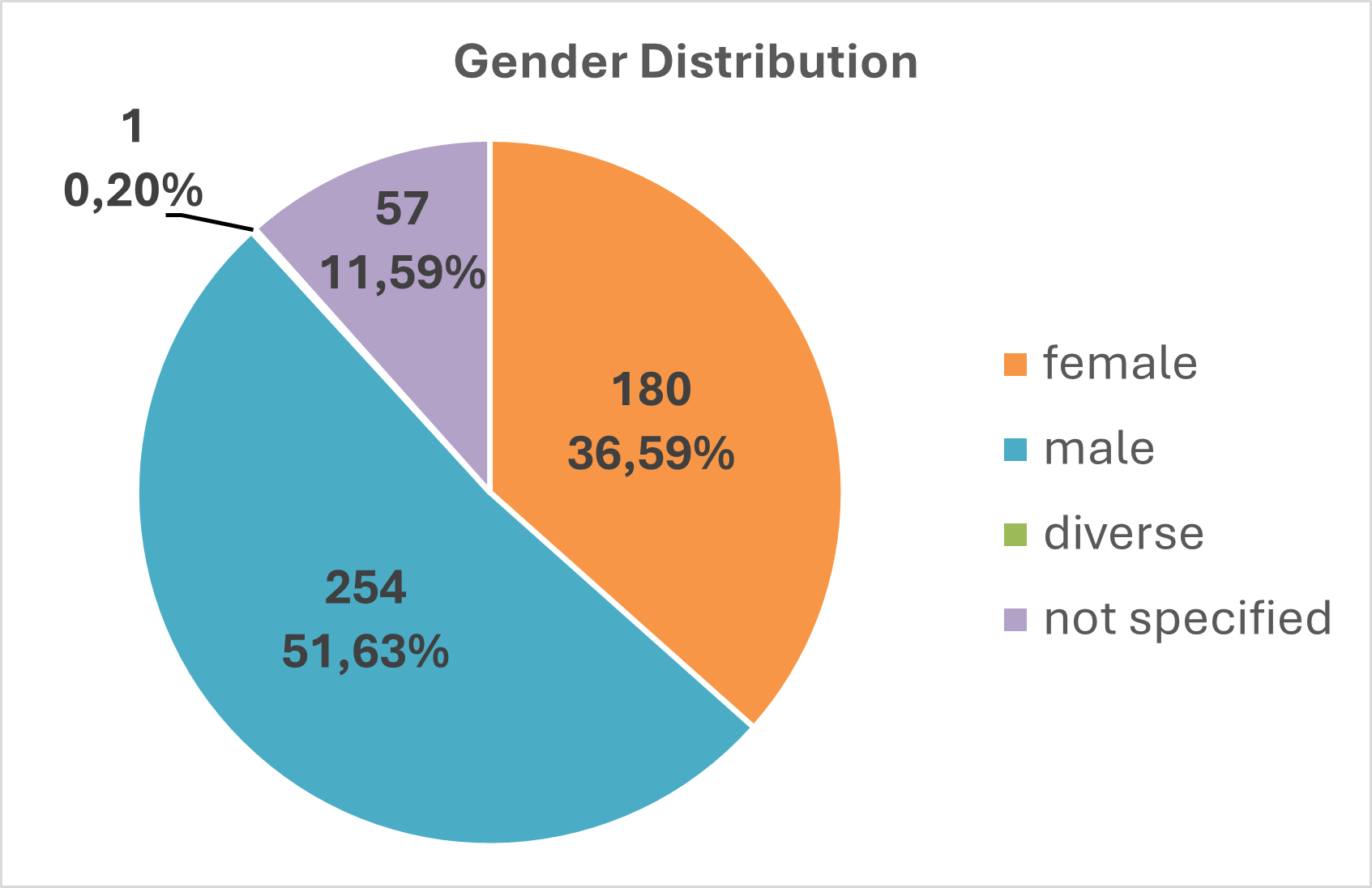}
\begin{center}
    (b)
\end{center}
\end{minipage}
\caption{(\textbf{a}) Age distribution of \gls{bez} probands; (\textbf{b}) Gender distribution of \gls{bez} probands }
\label{fig:bez_gender_distribution}
\end{figure}

\noindent With the above acquisition approach we recruited 492 participants, which have finished the enrollment process. The participant pool consists of 254 male, 180 female, 1 diverse and 57 not specified individuals (see Fig.~\ref{fig:bez_gender_distribution}). 428 have specified their birth date or year of birth. The average age of these probands is 30.6 ± 11.4. The oldest individual is 89 years old, the lowest specified age is 18. Two pairs of biological identical twins participated in the evaluations.

The evaluations took place three days a week in time slots of 2.5 hours. On average, around 96 participants took part in the evaluations per week and around 32 per evaluation day, whereby these figures can vary greatly with up to 146 visits per week depending on the time of year, among other things due to vacation time, lecture-free period.

Each participant provided data across several sessions on multiple experiment stations. These stations were arranged to facilitate smooth participant flow and allow for simultaneous recordings and minimizing waiting times. The collected datasets (see Table~\ref{tab:experiment_data}) include multiple biometric modalities, each with specific capture devices and recording conditions. At the already mentioned K13 photo studio, high-quality images from 13 perspectives of the subjects were recorded. In addition, \gls{abc} gates operated as at major national airports as well as an experimental gate and a stand-alone gate panel are in use. Long-term test series are of particular importance for these \gls{abc}-gates, as the effects of changing characteristics (skin, hairstyle, beard) pose a challenge, especially in view of the long validity of passports. At the fingerprint stations we alternated between optical and \gls{tft} based fingerprint scanners and fingerprints have been captured from both hands in flat and rolled mode.
Additionally the remission of skin is evaluated by capture of spectra in the visible and \gls{nir} range and the unique hardware \gls{pad} system LokiMk2 \cite{scheer_customizable_2024} station. At the thermal camera station, different camera systems captured thermal images of the subjects face. Another camera station used the \gls{tof} technology to capture 3D face data of subjects and is currently combined with \gls{rppg}. In addition to the long term tests, there were also individual sessions, so-called mass tests, in which special stations were set up and tested. These include the \gls{oct} experiment \cite{kirfel2022robust} where 3D fingerprints have been captured and an experiment with the focus on the back of the hand, where a camera and light system has been setup, to capture high-quality images of the back of the hand in different hand gestures. To evaluate the usability and human-technology interaction under realistic conditions, a dedicated study was conducted at the \gls{abc}-gates as part of a research project at the \gls{bez} \cite{paul2022usability}.

In addition to the real-world dataset collected through these test stations, a synthetic database was created to supplement the study. This synthetic dataset was generated using advanced character creation tools and game engines. MetaHumans \cite{epic_metahuman_2025}, Character Creator \cite{reallusion_charactercreator_2025} and the Blender Plugin HumanGenerator \cite{humangenerator_2025} have been used to create digital characters. MetaHumans and Character Creator have been used to compare real face data with synthetic face data in \cite{blumel_enhancing_2024}.

\begin{table}
\begin{center}
\caption{Experiment measurement data}
\label{tab:experiment_data}
\begin{tabular}{lrrr}
\hline
Experiment & Quantity & First meas. & Last meas.\\
\hline
ABC-Gates & 32908 & 2022-06-02 & 2025-02-18\\
K13 & 2779 & 2022-06-07 & 2025-02-18\\
Optical finger & 116920 & 2022-06-21 & 2025-02-11\\
TFT finger & 28903 & 2022-08-17 & 2025-02-11\\
Thermal cameras & 3340 & 2022-07-13 & 2023-03-09\\
TOF camera & 1710 & 2022-08-10 & 2023-03-09\\
TOF\_rPPG & 4237 & 2023-10-25 & 2025-02-18\\
Spectrometer & 24727 & 2022-08-17 & 2025-02-18\\
LokiMk2 & 22880 & 2023-03-23 & 2025-02-18\\
OCT & 201 & 2021-10-01 & 2021-11-08\\
Back of hand & 1037 & 2022-10-12 & 2022-11-09\\
\hline
\end{tabular}
\end{center}
\end{table}

\subsection{Analysis}
\noindent For the processing and analysis of these biometric data, a comprehensive infrastructure of hardware and software tools is available. The collected data is processed on a dedicated analysis server, which grants access to the data recorded in the \gls{bez} only in compliance with data protection and IT security requirements and ensures high computational performance for large-scale biometric evaluations \cite{von_twickel_konzept_2020}. A web-based, modular analysis framework facilitates the systematic examination of biometric characteristics, supporting various evaluation pipelines, including preprocessing and matching without direct access to the underlying biometric data. Operators can use the comprehensive data labeling scheme and corresponding filter options to extract the desired sub datasets relevant for the respective investigation. Due to the modular and flexible design of the framework, selected databases, pre- and post-processing components as well as examined biometric algorithm can be easily changed as required. Moreover, specialized modules managed and developed by individual operators can expand the framework. New biometric algorithms for comparison, presentation attack detection, morphing detection, deepfake detection or other, e.g. submitted by external vendors, can be flexibly integrated and investigated in a trustful environment.

The focus of this study is limited to the long-term effects of face images in regards to recognition accuracy, which is examined across different \gls{fr} algorithms. Face comparison scores have been obtained from \gls{cots} algorithmns, here called COTS1 and COTS2, as well as for the open-source algorithms ArcFace \cite{deng_arcface_2018}, AdaFace \cite{kim_adaface_2022} and DLIB \gls{fr} \cite{e_king_dlib-ml_2009, parkhi_deep_2015} to allow for comparison in-between different software. 

\subsection{Definitions}
\noindent The following terminology corresponding to the ISO 30107 \cite{international_organization_for_standardization_information_2023}, ISO/IEC 2382-37 \cite{international_organization_for_standardization_information_2022} and the BSI TR-03166 \cite{bsi_bsi_2024} is defined for this study. The term “comparison score” refers to an evaluation of the comparison of two facial images by a \gls{fr} algorithm and is therefore a way of estimating the similarity of two faces. If the score exceeds a threshold value set for the respective application, it is assumed that both images originate from the same person (“mated”). In the \gls{fr} algorithms used, the comparison score is specified in an interval between "0" and "1", with "1" corresponding to the best possible match between the two facial images. 

Furthermore, the \gls{far} denotes the percentage of falsely accepted unauthorized subjects. This error rate is important for assessing the security of a biometric procedure. The lower the \gls{far}, the more secure the procedure is. For the evaluation usability, the \gls{frr} is used analogously. The \gls{frr} is the rate of wrongly rejected authorized persons. The \gls{far} and \gls{frr} are often plotted as a function of the acceptance or threshold value. The \gls{far} decreases as the threshold value increases because the system becomes more restrictive and grants access to fewer people. For the same reason, the \gls{frr} increases. These two error rates depend on the threshold value and correlate with each other. It is not possible to minimize both. Therefore, when selecting the threshold value a trade-off must be made between safety (low \gls{far}) and usability (low \gls{frr}). The intersection of the \gls{far} curve with the \gls{frr} curve is called the \gls{eer}. When selecting the threshold value, the course of the \gls{far} and \gls{frr} curves around the \gls{eer} point should also be considered. A wide “valley” around this point would be desirable.

\begin{equation}
\label{far_ex1}
\mathit{FAR} = \frac{number\ of\ false\ acceptances}{number\  of\ unauthorized\ access\ attempts}
\end{equation}

\begin{equation}
\label{frr_ex1}
\mathit{FRR} = \frac{number\ of\ false\ rejections}{number\  of\ authorized\ access\ attempts}
\end{equation}

The \gls{fta} rate defines the proportion of cases in which no template could be generated from the given input image. In these cases, no comparison value can be determined.

Additionally, the \gls{fmr} is defined as the percentage of biometric comparisons of different identities that incorrectly result in a match. The \gls{fnmr} is accordingly the proportion of mated comparisons with a comparison score above the recognition threshold. 

\begin{equation}
\label{fmr_ex1}
\mathit{FMR} = \frac{number\ of\ non\_mated\ scores\ above\ threshold}{number\ of\  non\_mated\ comparisons}
\end{equation}

\begin{equation}
\label{fnmr_ex1}
\mathit{FNMR} = \frac{number\ of\ mated\ scores\ below\ threshold}{number\ of\  mated\ comparisons}
\end{equation}

Thus, \gls{fmr} and \gls{fnmr} are related to the \gls{far} and \gls{frr}, but refer to only the completed comparisons without regard of \gls{fta} cases. This relationship between \gls{fmr} and \gls{far} as well as \gls{fnmr} and \gls{frr} can therefore be defined as follows.

\begin{equation}
\label{far_fmr_ex1}
\mathit{FAR} = (1-\mathit{FTA}) \cdot \mathit{FMR}
\end{equation}

\begin{equation}
\label{frr_fnmr_ex1}
\mathit{FRR} = \mathit{FTA} + (1-\mathit{FTA}) \cdot \mathit{FNMR}
\end{equation}

To evaluate the temporal dynamics of biometric recognition performance across the full dataset, we define the delta score $\Delta s$ for each mated comparison as the difference between the similarity score at time $t$ and the baseline score of the same subject at the initial enrollment:

\begin{equation}
\label{ds_ex1}
\Delta s = s_{i0} - s_{ij}
\end{equation}

where $s_{i0}$ is the similarity score between the subject’s enrollment image and itself, and $s_{ij}$ is the score between the enrollment image and a follow-up image taken at time $t_j$. This transformation standardizes scores relative to each subject’s initial appearance, to allow a comparison across individuals.

The dataset is filtered to include only genuine comparisons within a defined time range. Individuals with only one recorded image are excluded, ensuring the temporal component is meaningful. For each subject $i$, the initial timestamp is determined, and the relative number of days since the initial image is computed:

\begin{equation}
\label{t_ex1}
\tilde{t}_{tj} = t_{ij} - t_{i0}
\end{equation}

A single dataset is then created, where each row corresponds to a comparison between a later probe and the original enrollment image for the same individual. The analysis has been applied to the four mentioned face recognition algorithms, each producing a variant of the delta score dataset. 

The delta score $\Delta s$ is modeled as a linear function of the relative time $\tilde{t}$, aggregated over the entire subject population:

\begin{equation}
\label{ds_ex2}
\Delta{s} = \beta_0 + \beta_1 * \tilde{t} + \varepsilon
\end{equation}

Here, $\beta_1$ captures the average rate of degradation in recognition score over time. The assumption is that all subjects follow a common temporal trend, even though specific individual trajectories may vary. The regression is performed over the entire set of mated comparisons, without nested modeling per individual.

\FloatBarrier
\section{Results}
\subsection{Data}
\begin{table}
\begin{center}
\caption{Experiment measurement data}
\label{tab:datasets_k13}
\begin{tabular}{lr}
\hline
Dataset & Number of Measurements\\
\hline
\textbf{D\_0: all K13 Data} & \textbf{2779}\\
D\_1.1: relevant measurements & 2767\\
D\_1.2: not relevant & 12\\
\textbf{D\_1.1: relevant measurements} & \textbf{2767}\\
D\_2.1: neutral face & 2480\\
D\_2.2: facial expressions & 287\\
\textbf{D\_2.1: neutral face} & \textbf{2480}\\
D\_3.1: no glasses & 2476\\
D\_3.2: glasses & 4\\
\hline
\end{tabular}
\end{center}
\end{table}

\noindent For the period under consideration (2022-06-07 - 2025-02-25) the K13 dataset can be summarized by the data subsets listed in Table~\ref{tab:datasets_k13}. Thus, from the overall $2779$ recordings, those $2476$ relevant measurements with neutral face and without glasses were taken into account. As described in section~\ref{sec:methodology} every recoding consists of 13 image from the different camera angles.

\subsection{Biometric Performance}
\begin{figure}[!t]
\centering
\includegraphics[width=3in]{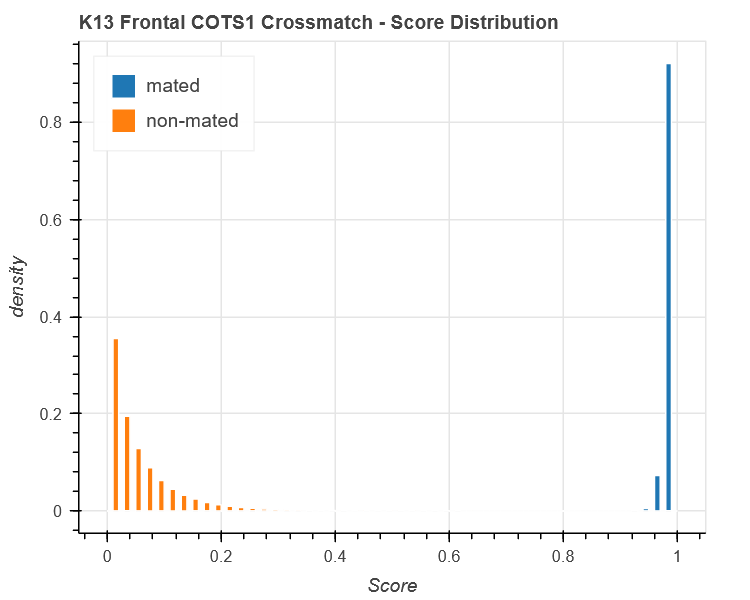}
\caption{K13 frontal images COTS1 score distributions for mated and non-mated comparisons. A sharp separation between mated and non-mated comparison scores, except of only $24$ outliers for biological identical twins, can be found.}
\label{fig:k13_cots1_score_distribution}
\end{figure}

\begin{figure}[!t]
\centering
\includegraphics[width=3in]{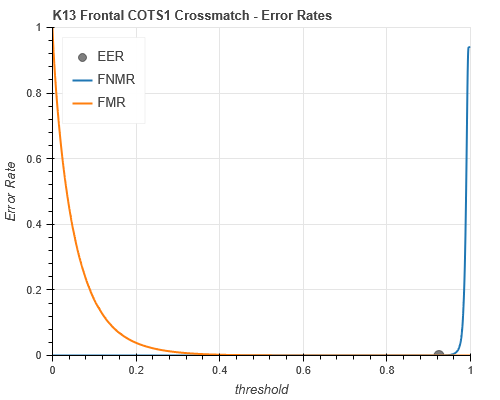}
\caption{K13 frontal images COTS1 score error rates based on different thresholds. With increasing threshold values the \gls{fmr} decreases because the system becomes more restrictive and grants access to fewer people. On the other hand, the \gls{fnmr} increases}
\label{fig:k13_cots1_error_rates}
\end{figure}

\noindent As a first evaluation of the recorded data and a benchmark for the longitudinal studies we investigated the performance of \gls{cots} and open-source \cite{deng_arcface_2018, e_king_dlib-ml_2009, parkhi_deep_2015, kim_adaface_2022} \gls{fr} algorithms on the K13 dataset. Initially, using only the frontal images of camera 5 for the D\_3.1 dataset (see Table~\ref{tab:datasets_k13}) results in the comparison score distribution shown in Fig.~\ref{fig:k13_cots1_score_distribution} and the corresponding statistics summarized in Tables \ref{tab:k13_data_cots1} and \ref{tab:k13_scores_and_eer_cots1}. As expected of a high quality data set in terms of the record image quality and the data labeling, a sharp separation between mated and non-mated comparison scores can be found. The \gls{eer} of $ 3.94 \cdot 10^{-6}$  is achieved for a threshold of $0.92$ here (see Fig.~\ref{fig:k13_cots1_error_rates}). However, for a real application scenario the threshold would be chosen quite lower to reduce false rejections due to potentially variations and lower qualities than these highly controlled, high quality K13 images. A threshold of $T=0.755$ seems to be more realistic as it is the lower boundary that would lead to the same error rates and the same number of outliers: still no false rejections and only $24$ false acceptances occur. A closer look to these $24$ outlier comparisons shows, that the compared subjects are biological identical twins, which look very similar even for a human operator. The mean mated score for these twins is $0.9938 \pm 0.0047$, whereas the mean non-mated score is $0.9640 \pm 0.0067$.
Therefore, differentiability based on this comparison algorithm seems theoretically possible in this special case with the high quality K13 data with minimal variations.

\begin{table}
\centering
\caption{K13 Frontal COTS1 Data}
\label{tab:k13_data_cots1}
\begin{tabular}{c|c|c|c|c}
\hline
\multicolumn{2}{l}{\# of images} & \multicolumn{2}{l}{\# of comparisons} & \multirow{1}{*}{FTA Rate} \\
probes        & candidates       & mated           & non-mated           &                           \\
$2476$          & $2476$             & $41600$           & $6088976$             & $0.0\%$                     \\
\hline
\end{tabular}
\end{table}

\begin{table}
\centering
\caption{K13 Frontal COTS1 Scores and EER}
\label{tab:k13_scores_and_eer_cots1}
\begin{tabular}{llr}
\hline
\multirow{2}{*}{Avg Score}  & mated     & $0.9885 \pm 0.0064$    \\
                            & non-mated & $0.0554 \pm 0.0634$    \\
\multicolumn{2}{l}{EER}                 & $3.94 \cdot 10^{-6}$   \\
\multicolumn{2}{l}{EER threshold}       & $0.9245$               \\
\multirow{2}{*}{Outliers}   & mated     & $0$                    \\
                            & non-mated & $24$                   \\
\hline
\end{tabular}
\end{table}

Evaluating additional comparison algorithms leads to similar results as found in Table~\ref{tab:k13_algorithms_eer_and_fta}. All tested algorithms confirm a clear separation between mated and non-mated data. Only the DLIB face recognition \cite{e_king_dlib-ml_2009, parkhi_deep_2015} achieved a slightly higher error rate and more outliers, which can be traced back to lower biometric performance. The high \gls{fta} rate in which the algorithm was unable to create a template from the input images is also noticeable. It was observed that the very high resolution of the K13 data sometimes leads to problems and that reducing the resolution can improve face detection.

\begin{table}
\centering
\caption{K13 Algorithms EER and FTA}
\label{tab:k13_algorithms_eer_and_fta}
\begin{tabular}{ccccc}
\hline
\multirow{2}{*}{Algorithm}   & \multirow{2}{*}{EER} & \multirow{2}{*}{FTA Rate}      & \multicolumn{2}{c}{Outliers} \\
            &                           &               & mated     & non-mated         \\
COTS1       & $3.94 \cdot 10^{-6}$      & $0.0\%$       & 0         & 24                \\
COTS2       & $3.94 \cdot 10^{-6}$      & $0.0\%$       & 0         & 24                \\
ArcFace     & $3.94 \cdot 10^{-6}$      & $0.08\%$      & 0         & 24                \\
AdaFace     & $3.94 \cdot 10^{-6}$      & $0.0\%$       & 0         & 24                \\
DLIB        & $2.44 \cdot 10^{-4}$      & $1.61\%$      & 10         & 1462              \\
\hline
\end{tabular}
\end{table}

These findings demonstrate that, except for biological identical twins, all subjects can successfully be identified by their K13 frontal face images using performant \gls{cots} or open-source comparison algorithms. No miss-labeled data could be found in the database, which highlights the high quality dataset acquired in the \gls{bez} evaluations.

\subsection{Angle Analysis}
\begin{figure}[!t]
\centering
\includegraphics[width=3in]{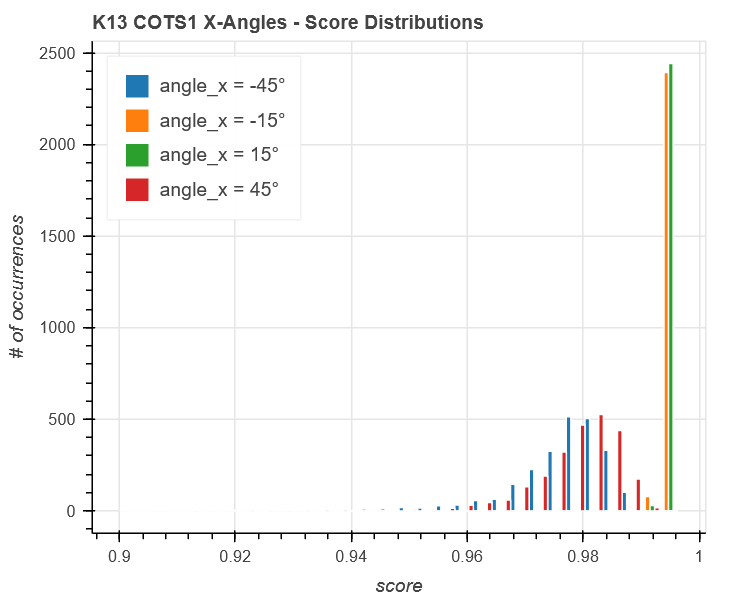}
\caption{K13 images with different horizontal record angles compared to frontal image of same record using COTS1. With higher x-angles, the comparison scores slightly decrease.
}
\label{fig:k13_cots1_x_angle_scores}
\end{figure}

\begin{figure}[!t]
\centering
\includegraphics[width=3in]{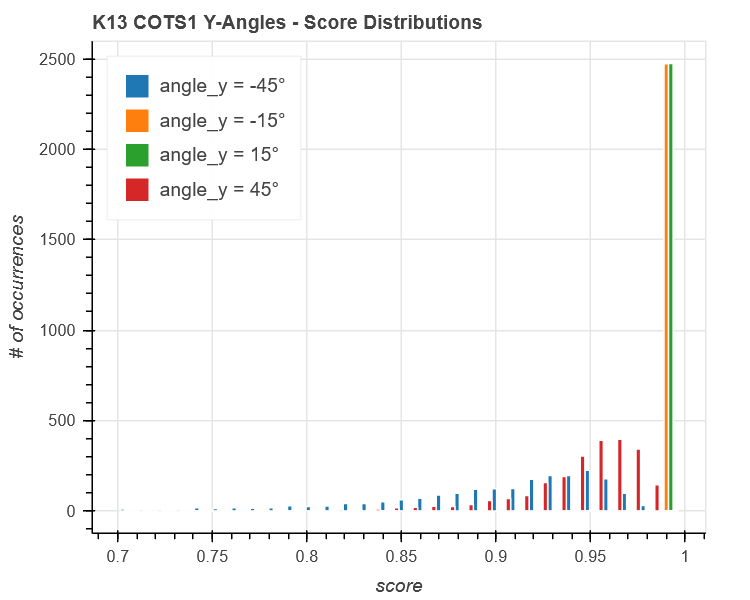}
\caption{K13 images with different vertical record angles compared to frontal image of same record using COTS1. With higher y-angles, the comparison scores decrease significantly and partially fall below the threshold T=0.755.
}
\label{fig:k13_cots1_y_angle_scores}
\end{figure}

\noindent In the next study, the full frontal K13 images of camera 5 (see Fig.~\ref{fig:k13_experiment_setup}) was compared to the additional cameras, whereby only comparisons to the different angles of the same recordings were considered. By this method only the influence of the angle is decisive and not the variance of different images of the same person. We define the x-angle as the rotation around the subject on a horizontal plane (left-right movement) and y-angles as the vertical elevation respectively (up-down movement). Figures \ref{fig:k13_cots1_x_angle_scores} and \ref{fig:k13_cots1_y_angle_scores} as well as Tables \ref{tab:k13_x_angle_scores} and \ref{tab:k13_y_angle_scores} show the results for various x- (horizontal) and y-angles (vertical), whereas the respective other angle is 0. Thus, influence of horizontal and vertical angles on the comparison scores can be investigated separately. Angle 0 would accordingly be a comparison to the same image and is therefore not plotted here.

The plotted distribution and detailed results show, that the higher the angle the lower the comparison score. This is the case for both x- and y-angles, but y-angles (vertical) have significantly greater effects. In some cases the increased record angle leads to a \gls{fta} and the comparison algorithm is not able to enroll a biometric template. Tables also list the resulting \gls{frr} the for the above defined threshold $T=0.755$, including the \gls{fta} cases. This confirms that higher vertical record angles remarkably increase false rejections, particularly due to higher FTA rate. Furthermore, a symmetry can essentially be recognized, i.e. positive and negative angles exhibit a similar effect, especially for horizontal x-angles. For the vertical y-angles, the negative angle and therefore a perspective from below has a greater effect with increased errors.

\begin{table*}
\centering
\caption{K13 x-angle COTS1 scores and error rates}
\label{tab:k13_x_angle_scores}
\begin{tabular}{lllll}
                    & $\alpha_{x}$ = $15$°   & $\alpha_{x}$ = $-15$°  & $\alpha_{x}$ = $45$°   & $\alpha_{x}$ = $-45$°  \\
\hline
 Mean Score mated   & $0.9950$          & $0.9947$          & $0.9782$          & $0.9756$          \\
 FTA Rate           & $0.04\%$          & $0.20\%$          & $1.01\%$          & $1.25\%$          \\
 \# below $T=0.755$   & $0$               & $0$               & $0$               & $1$               \\
 FRR for $T=0.755$    & $0.04\%$          & $0.20\%$          & $1.01\%$          & $1.29\%$          \\
\hline
\end{tabular}
\end{table*}

\begin{table*}
\centering
\caption{K13 y-angle COTS1 scores and error rates}
\label{tab:k13_y_angle_scores}
\begin{tabular}{lllll}
                    & $\alpha_{y}$ = $15$°   & $\alpha_{y}$ = $-15$°  & $\alpha_{y}$ = $45$°   & $\alpha_{y}$ = $-45$°  \\
\hline
 Mean Score mated   & $0.9946$          & $0.9946$          & $0.9385$          & $0.8912$          \\
 FTA Rate           & $0\%$             & $0.04\%$          & $6.38\%$          & $11.55\%$         \\
 \# below $T=0.755$   & $0$               & $0$               & $23$              & $138$             \\
 FRR for $T=0.755$    & $0\%$             & $0.04\%$          & $7.31\%$          & $17.12\%$         \\
\hline
\end{tabular}
\end{table*}

\subsection{Longterm Analysis}
\begin{figure}[!t]
\centering
\includegraphics[width=3.5in]{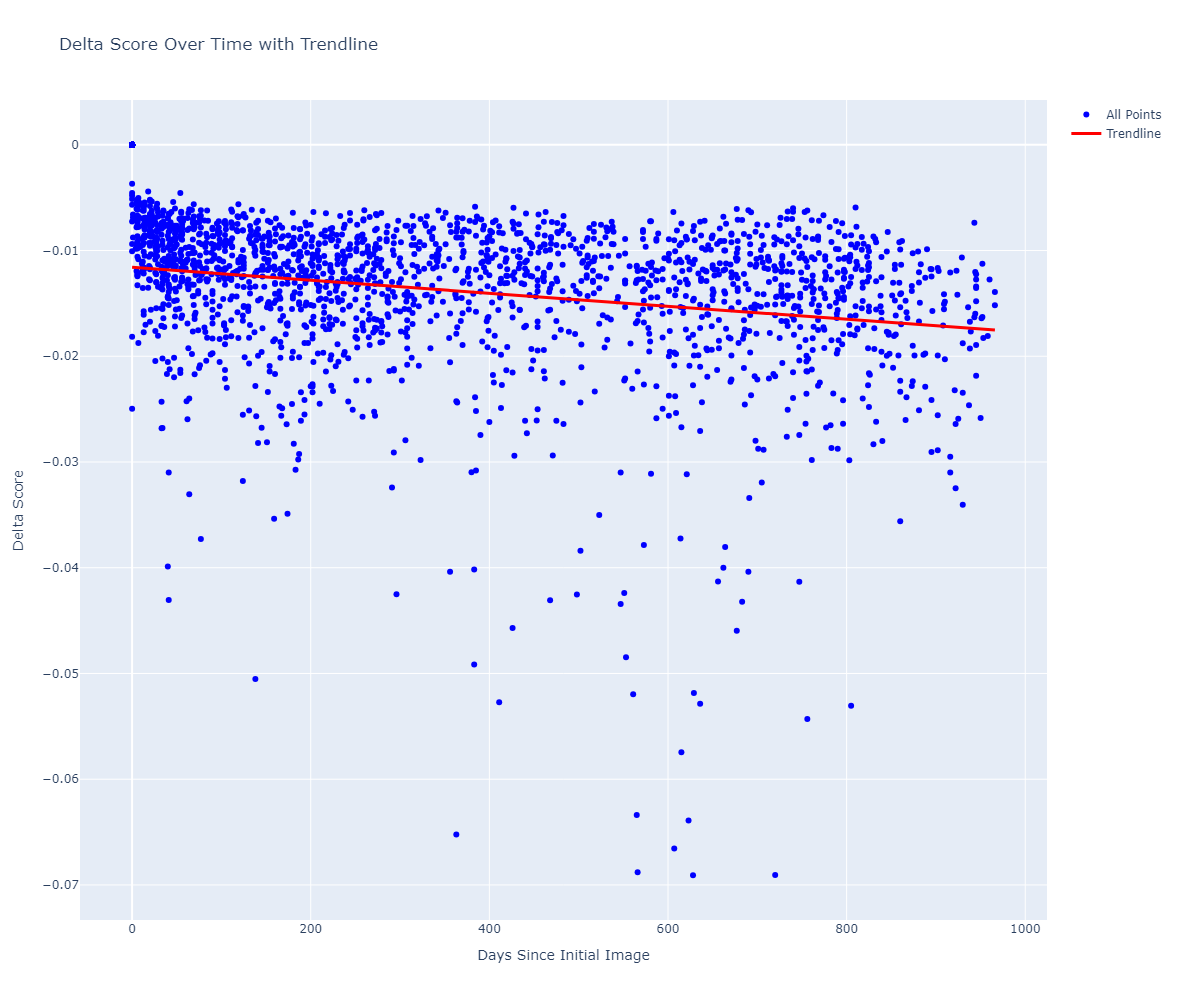}
\caption{Frontal K13 image comparison with the same persons initial frontal image using COTS1. A global trendline (red) indicates a slight decrease over the increasing time. A comparison of the initial frontal image with the same initial frontal image results in a delta score of 0.
}
\label{fig:k13_cots1_delta_scores}
\end{figure}

\noindent To analyze the temporal stability of facial recognition performance, delta scores were plotted against the number of days since the initial image acquisition for all participants using the \gls{cots} and open-source algorithms. As shown in Fig. \ref{fig:k13_cots1_delta_scores}, each point represents the delta score for an individual genuine comparison relative to the first capture.

The scatter plot reveals a downward trend over time, which is further supported by the fitted linear regression line. This indicates a gradual decline in similarity scores, suggesting that the facial recognition performance of the COTS1 algorithm deteriorates slightly as the time gap between image acquisitions increases.

\begin{figure}[!t]
\centering
\includegraphics[width=3in]{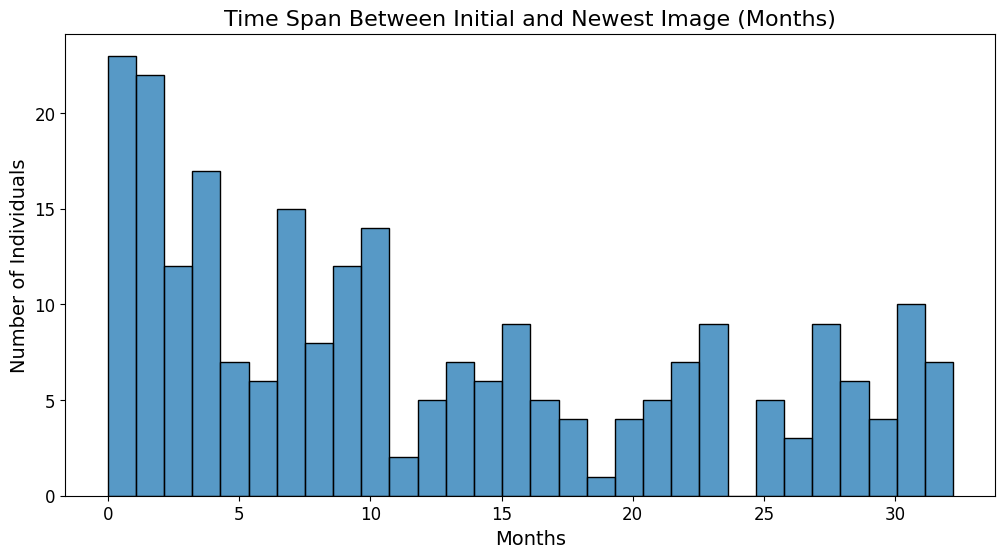}
\caption{Aggregated amount of subjects with the time-span from the initial and the newest captured image in months. The amount of subjects is high in the initial months and decreases with higher values.
}
\label{fig:k13_timespan_initial_newest_image}
\end{figure}

While individual variations exist and intra-day similarity scores fluctuate stronger than inter-day similarity scores, the general trend suggests a consistent and measurable temporal drift in facial features. The distribution is densest in the early time intervals, as can be seen in Fig. \ref{fig:k13_timespan_initial_newest_image}. Here delta scores tend to cluster closer to zero, whereas longer intervals are associated with more negative delta scores. This aligns with expectations from facial biometric systems, where changes in appearance due to aging or other temporal factors can reduce match confidence over time.

\begin{figure}[!t]
\centering
\includegraphics[width=\linewidth]{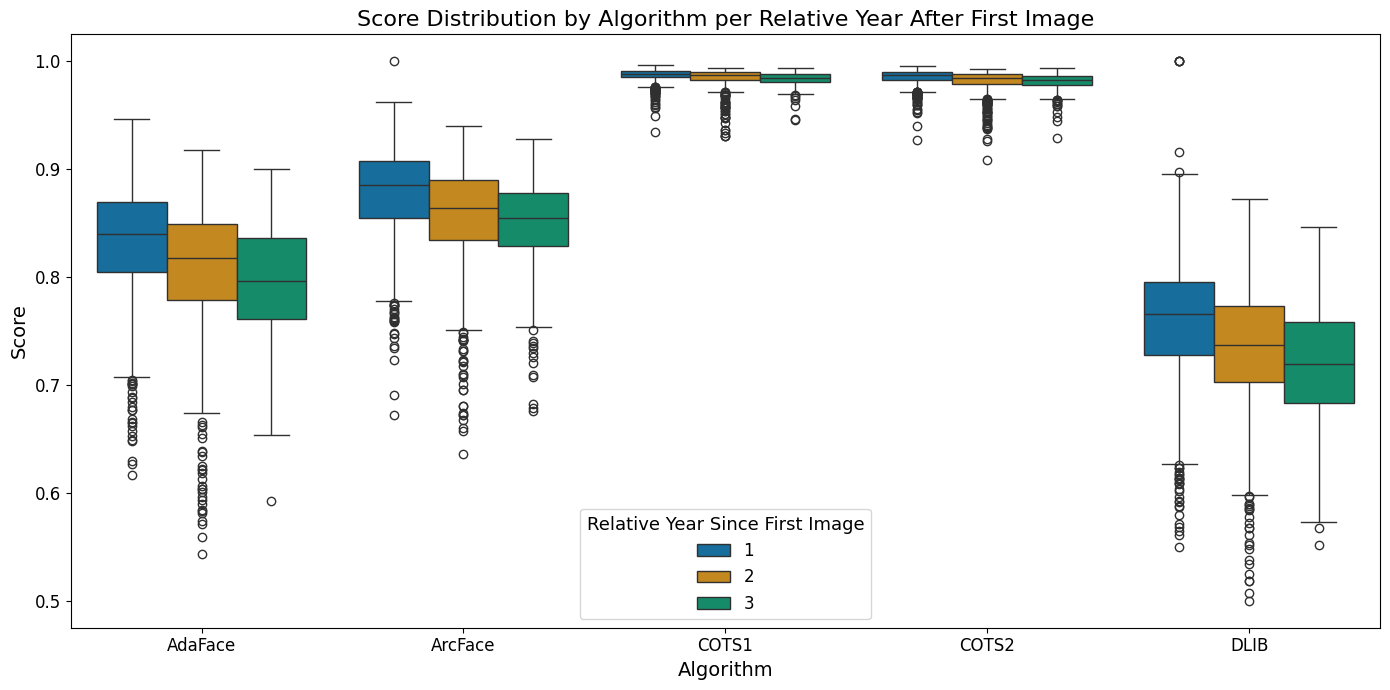}
\caption{Boxplot of the score distribution per algorithm and year. 
}
\label{fig:boxplot_longterm_score_dist_algorithm_per_year}
\end{figure}

A similar analysis was conducted for the remaining four facial recognition algorithms as seen in Fig. \ref{fig:boxplot_longterm_score_dist_algorithm_per_year}. Across all algorithms, a general decline in delta scores over time was observed. However, the magnitude of this decline varied significantly between algorithms. The effect on the delta score is not directly comparable between the algorithms, as their functionality differs from each other. Therefore for each algorithm a separate threshold has been calculated. The open-source algorithms exhibit a noticeably stronger negative trend in their delta scores compared to the COTS algorithms. This suggests that these algorithms may be more sensitive to temporal variation and possibly less robust in handling aging-related facial changes or other longitudinal appearance shifts.

\begin{table}
\centering
\caption{cots1 longitudinal statistics and delta scores}
\label{tab:cots1_statistics}
\begin{tabular}{lrrr}
Year    & 1.    & 2.  & 3. \\
\hline
Count   & 1333      & 475       & 264     \\
Mean    & -0.0093   & -0.0149   & -0.0155 \\
Std     & 0.0074    & 0.0078    & 0.0068  \\
Min     & -0.0652   & -0.0691   & -0.0543 \\
25\%    & -0.0132   & -0.0173   & -0.0180 \\
Median  & -0.0094   & -0.0128   & -0.0143 \\
75\%    & 0.0000    & -0.0100   & -0.0108 \\
Max     & 0.0       & -0.0059   & -0.0059 \\
\hline
\end{tabular}
\end{table}

In addition to the observable period this study is limited to, the linear trend allow for a cautious extrapolation of future delta scores. This projected decline, although hypothetical, suggests a critical threshold where the biometric similarity score may fall below operational match thresholds, especially in less robust algorithms. Such long-term deterioration emphasizes the necessity for periodic re-enrollment.

\section{Discussion}
\noindent The results of this longitudinal analysis confirm that facial recognition performance is affected by temporal variation. Across all four algorithms tested, a general decline in similarity was observed over time. While the trend was present in all algorithms, the magnitude of degradation varied significantly. The \gls{cots} algorithms demonstrated more stable behavior over time, while open-source alternatives showed a steeper decline in performance.

These findings suggest that algorithmic robustness plays a crucial role in maintaining reliable identity verification over extended periods. Moreover, the relatively dense data points and consistent time intervals achieved through the controlled study setting at the \gls{bez} provide a high level of temporal resolution. This stands in contrast to other studies, which often span several years with irregular recording intervals and lack consistent environmental conditions \cite{best-rowden_longitudinal_2018}.

Extrapolating the linear trendline further suggests a future point at which recognition performance could fall below acceptable thresholds, particularly in non-adaptive systems. 

The dataset acquired through this long-term study offers a unique opportunity for biometric research, particularly in assessing the stability and variance of facial biometric data over time. Our findings reveal that temporal changes in facial comparison scores over 2.5 years are minimal in \gls{cots} algorithms, with inter-individual score distances remaining significantly larger than intra-individual temporal variations. This underscores the reliability of facial biometrics as a stable identifier over time.

Ethical considerations were central to the study design. Advanced data protection measures (anonymization and pseudonymization of acquired data, storage on disconnected local servers, strict appliance of \gls{gdpr}-guidelines) and an automated kiosk-based registration system ensured the privacy and security of participants. This approach serves as a scalable and privacy-respecting model for future biometric research.

Future work will focus on improving synthetic character generation technologies to better emulate the diversity and complexity of real-world biometric data, thereby enhancing their applicability in large-scale biometric system development.

\section{Conclusion}
\noindent In this work, we have presented a comprehensive controlled longitudinal evaluation of facial biometrics, spanning nearly three years and over 2476 high-quality frontal images. The controlled \gls{bez} setup allowed us to evaluate temporal effects in frontal face images. The resulting dataset exhibits exceptional integrity, evidenced by the near-zero \gls{fta} rates and the sharp separation of mated versus non-mated score distributions across both \gls{cots} and leading open-source algorithms.

Our analysis reveals a clear, statistically significant downward trend in genuine match scores as the time interval since enrollment increases. \gls{cots} systems demonstrate remarkable resilience, with only marginal score degradation over 2.5 years, whereas open-source models are more susceptible to temporal drift. Importantly, the observed rate of temporal score decline aligns closely with the European Union's current ID-card renewal guidelines, lending empirical support to existing policy.

These findings emphasize two crucial takeaways for biometric system designers and deployers: (1) longitudinal performance testing under consistent conditions is indispensable for understanding real-world system robustness; and (2) periodic re-enrollment or dynamic algorithm adaption enhances reliability over extended usage periods. The \gls{bez} infrastructure and dataset pave the way for further research into adaptive matching thresholds and the long-term utility of novel \gls{fr} architectures.

\section{Outlook}
\noindent While this study provides valuable insights into the temporal stability and variability of facial biometrics over a three-year period, several avenues for future research remain open. 

First, as the study at the \gls{bez} continues, the increasing volume of data per subject will enable more precise and statistically robust analyses. With a greater number of repeated measurements across longer time spans and a broader participant base, future evaluations can better capture individual and population-level trends in facial recognition performance. This will allow for more reliable estimations of temporal drift, potential thresholds for re-enrollment, and variations across demographic groups.

Second, extending the long-term analysis to include face images of lower quality, such as live-images obtained from \gls{abc}-gates. These images typically suffer from suboptimal lighting, pose variation and compression artifacts. Recognition systems often under-perform with these conditions.
In addition, the infrastructure at the \gls{bez} can be leveraged to expand the longitudinal analysis to other biometric modalities, such as fingerprints.

We are currently addressing these challenges through the continued collection of long-term biometric data and ongoing analyses. We believe these efforts will contribute to a more comprehensive understanding of biometrics and facilitate the development of robust, ethically sound systems for real-world applications.

\printbibliography
%%\bibliographystyle{IEEEbib}
%%\bibliography{bibliography}

\end{document}